# Mean of Means: Human Localization with Calibration-free and Unconstrained Camera Settings (extended version)

**Tian-Yi Zhang[1], Weng-Yu Zhang[1], Xulu Zhang[1], Xulu Zhang[1], Jiaxin Wu[1], Xiao-Yong Wei[2,1*], Jiannong Cao[1], Qing Li[1],**

[1]Hong Kong Polytechnic University, Hong Kong SAR
[2]Sichuan University
* Corresponding author: Xiao-Yong Wei (cswei@scu.edu.cn, x1wei@polyu.edu.hk)

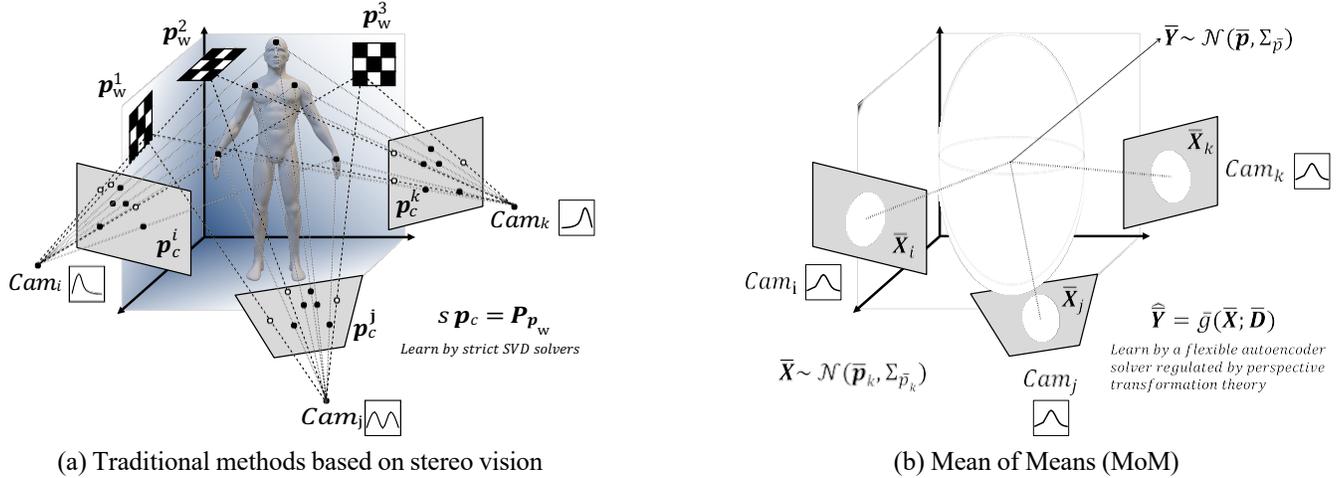

(a) Traditional methods based on stereo vision  (b) Mean of Means (MoM)

Figure 1: Comparison of traditional method and Mean of Means (MoM): We have transformed the learning goal from the point-based relation between the world and pixel coordinates to the relation of distributions of these two sets of coordinates.

## Abstract

Accurate human localization is crucial for various applications, especially in the Metaverse era. Existing high precision solutions rely on expensive, tag-dependent hardware, while vision-based methods offer a cheaper, tag-free alternative. However, current vision solutions based on stereo vision face limitations due to rigid perspective transformation principles and error propagation in multi-stage SVD solvers. These solutions also require multiple high-resolution cameras with strict setup constraints. To address these limitations, we propose a probabilistic approach that considers all points on the human body as observations generated by a distribution centered around the body's geometric center. This enables us to improve sampling significantly, increasing the number of samples for each point of interest from hundreds to billions. By modeling the relation between the means of the distributions of world coordinates and pixel coordinates, leveraging the Central Limit Theorem, we ensure normality and facilitate the learning process. Experimental results demonstrate human localization accuracy of 96% within a $0.3m$ range and nearly 100% accuracy within a $0.5m$ range, achieved at a low cost of only 10 USD using two web cameras with a resolution of 640×480 pixels. The dataset and source code can be accessed at https://github.com/open_upon_acceptance.

## Introduction

Human localization plays a crucial role in various location-based applications, including AR/VR, indoor navigation, fitness and health tracking, surveillance and security, and sports analysis. The growing prominence of Metaverses and Digital Twins has further emphasized its significance (Wang et al. 2022; Mihai et al. 2022; Zhou et al. 2023; Li et al. 2023). For instance, in Metaverse, accurately determining user's location is essential to generate a corresponding virtual representation within the Metaverse. In healthcare applications of Digital Twins, tracking the user's location is crucial for behavior analysis as locations are closely linked to user activities (e.g., the user's water intake behavior can be inferred from their frequency of standing in front of the water dispenser). Current human localization solutions primarily rely on hardware-based approaches such as UWB (Poulose and Han 2021), Bluetooth (Li et al. 2020), WiFi (Pu et al. 2023), and LiDAR (Dai et al. 2022). However, this hardware dependency limits their applicability in broader scenarios due to the associated costs for base stations. Additionally, in most hardware-based solutions, every user needs to carry a tag/device. Dependence on expensive tags not only adds to cost but also introduces inconvenience.

To eliminate the need of user tags, one may consider vision-based methods. However, existing vision solutions are far more expensive than hardware-based ones in two aspects. In terms of equipment, these solutions are primarily designed for film production or high-fidelity performance capture, necessitating high-resolution camera arrays with tightly controlled configurations, thus predominantly utilized in motion capture studios (Islam et al. 2020; Sinha et al. 2020). Consequently, these solutions are impractical for Metaverse or Digital Twins applications where a lightweight, easily deployable solution is essential. Furthermore, those solutions are technically expensive. With the focus on high-quality motion or 3D structure modeling, those methods are developed to model the location (i.e., world coordinate $\mathbf{p}_w$) of every point $\mathbf{p}$ on the target in related to its pixel coordinates $\mathbf{p}_c$ though a transformation matrix $\mathbf{P}$ as

$$s\mathbf{p}_c = \mathbf{P}\mathbf{p}_w, \quad (1)$$

where $s$ is a scale factor. The $\mathbf{P}$, encapsulating the intrinsic and extrinsic parameters, has to be estimated through a complex camera calibration process and the applications of SVD solvers multiple times. The estimation poses three challenges. Firstly, the collection of training pairs $(\mathbf{p}_c, \mathbf{p}_w)$'s is a tedious process of placing the chessboards at different angles and positions, but it can only collect samples on a limited scale of hundreds (Zhang 2000). Secondly, due to the highly deformable characteristics of the human body and the effects of perspective transformation, the coordinates of the target points follow unknown and heterogeneous distributions, which presents challenges for the commonly used singular value decomposition (SVD) solvers, in finding a global optimum (Golub and Reinsch 1971; Allen-Zhu and Li 2016; Ngo et al. 2005, 2008, 2010). Lastly, when considering the limited size of the training set and the heterogeneity of distributions, we will show in the Method that the expectation of the location prediction does not easily converge to that of the ground truth. This paper introduces a framework named Mean of Means (MoM) designed to tackle these challenges. This cost-effective solution leverages inexpensive web cameras and offers easy deployment by eliminating the need for intricate camera calibration. The core concept involves easing the stringent demand for precise world-pixel coordinate matches for all points by treating the entire human body as a distribution derived from its mean or geometric center. Through the adoption of this methodology, the focus transitions to understanding the relationship between the distributions of world coordinates and pixel coordinates, rather than analyzing them on a point-by-point basis. This shift improves the practicality and feasibility of the approach. The advantages are as follows.

**Large-Scale Sampling:** This relaxation significantly enhances the flexibility of data collection, as any sampled points from the body can be used as observations to estimate a single center. By sampling only 20 points from each body, we can theoretically collect trillions of training pairs, resulting in a considerably large training set (denoted as $\mathbf{D}$).

**Normality Consistency:** For the distribution heterogeneity, we propose modeling a relation between the world coordinate mean $\mathbf{Y}$ and corresponding pixel coordinate means $\mathbf{X}$ rather than focusing on point estimators. This leads to the learning problem of

$$\hat{\bar{\mathbf{Y}}} = \bar{g}(\bar{\mathbf{X}}; \bar{\mathbf{D}}). \quad (2)$$

As depicted in Fig. 1, this approach allows us to randomly combine observations of body points as estimators for the center $\mathbf{Y}$. The Central Limit Theorem (CLT) ensures that the sample mean will follow a normal distribution (i.e., $\bar{\mathbf{Y}} \sim \mathsf{N}(\bar{\mathbf{p}}, \Sigma_{\bar{p}})$) when the sample size is sufficiently large. The same principle applies to the pixel coordinates, making the means of pixel coordinates also follow normal distributions (i.e., $\bar{\mathbf{X}} \sim \mathsf{N}(\mathbf{p}_k, \Sigma_{\bar{p}_k})$). The consistent normality of these distributions enhances the feasibility of the model learning.

**Balance between Neural Implementation and Classical Theory:** as ensured by the Law of Iterated Expectations (LIE), the expectation of the new model $\mathsf{E}(\hat{\bar{\mathbf{Y}}})$ converges to the true expectation $\mathsf{E}(\mathbf{Y})$. We thus propose implementing MoM using an end-to-end autoencoder framework, where an encoder learns a neural mapping function $\bar{g}(\cdot)$ directly, replacing the multi-stage SVD solvers utilized in conventional methods. This reduces the risk of error propagation. The encoder also satisfies the conditions outlined in (Brutzkus and Globerson 2017), which state that a CNN with non-overlapping convolutions can achieve a global optimal solution. Moreover, to address the potential problem of overfitting in the neural implementation of the encoder, we introduce a decoder that follows the well-established perspective transform. The collaboration between the proposed encoder and decoder ensures a balance between the capabilities of neural networks and the established perspective transformation theory.

# Related Work

Human localization plays a crucial role in numerous applications, including 3D reconstruction (Andriluka et al. 2014; Song et al. 2021; Guzov et al. 2021), healthcare (Bharadwaj et al. 2017; Bibbo`, Carotenuto, and Della Corte 2022), and sports analysis (Ridolfi et al. 2018). In this section, we can only provide a concise overview of related work pertaining to our method due to space limitations. For a more comprehensive understanding, we recommend surveys in (Morar et al. 2020; Zafari, Gkelias, and Leung 2019; Yang, Cabani, and Chafouk 2021; Bibbo`, Carotenuto, and Della Corte 2022). There are two groups of methods that have been explored: hardware-based and vision-based approaches.

## Hardware-based Methods

This group revolves around signal-based techniques, in which three or more anchor or base stations are installed at fixed positions, which transmit and receive signals to a tag or device carried by the user. The position of a tag can be estimated using triangulation based on the time difference of arrival (TDOA) from all anchors/bases. The methods within this group differ primarily in the signals utilized, such as WIFI (Wang et al. 2016; Pu et al. 2023), Bluetooth (Kriz et al. 2016; Li et al. 2020), RFID (Ruan et al. 2018; Ma et al. 2020), and UWB (Cheng and Zhou 2019; Poulose and Han

2021). Hybrid methods have also been developed, which combine different hardware components to achieve more reliable performance (Monica and Bergenti 2019). Hardware-based methods can offer high precision (Wu 2022). However, these methods are still limited by the expensive cost of hardware, the complexity of setup, and the dependence on the specific tag/device.

### Vision-based Methods

Vision-based methods offer a more flexible setup (Morar et al. 2020) because of their tag-free nature. This category of methods has been extensively explored and encompasses various sub-categories. Due to space limitations, we will briefly introduce the following two categories.

**Stereo Vision Methods**: These methods follow a two-stage pipeline (Morar et al. 2020; Wu et al. 2022), consisting of a Perspective-n-Point (PnP) problem is solved to determine the transformation matrix (Sun et al. 2019b; Cosma, Radoi, and Radu 2019), and application of the triangulation principle using utilizing SVD solvers. The theory has been well-established (Zhang 2000) and widely applied in various tasks. However, previous applications have demonstrated that estimating world coordinates for all points while adhering to the principle is a highly rigid requirement as they rely on several assumptions, such as well-calibrated cameras (Yang et al. 2023; Wu et al. 2022; Jain et al. 2018) and strict constraints on camera positions and angles (Islam et al. 2020). Bundle adjustment is also frequently employed to further refine the transformation matrix and predicted world coordinates (Triggs et al. 2000; Zach 2014; Wei and Yang 2012). While stereo vision methods are commonly used for tasks like 3D reconstruction (Schonberger and Frahm 2016; Do and Nguyen 2019; Huang, Hu, and Zhang 2012) and pose estimation (Kang et al. 2023; Zhou et al. 2022; Li et al. 2023), our method is closely related to this category, as we apply the same theory to a different task. In addition, our method relaxes the requirement of estimating all points simultaneously and eliminates the need for camera calibration and strict installation constraints.

**Depth Estimation Methods**: The primary objective of this category is to estimate depth which is indirectly related to locations (Zhou et al. 2017; Bhoi 2019). Depth estimation can be achieved by leveraging motion or structural relationships among points on the target and applying principles derived from stereo vision theory to estimate scaled distances (Rajasegaran et al. 2022; Takacs, Vincze, and Richter 2020; Zheng et al. 2023; Wei and Yang 2011). In recent times, depth cameras have played a significant role in advancing depth estimation methods (Bhoi 2019; Masoumian et al. 2022). A popular approach involves training a depth estimator using paired images and depth maps (Ranftl et al. 2022; Zhang and Funkhouser 2018). However, this approach becomes reliant on the availability of depth cameras for training, which can be both costly and susceptible to interference. Depth cameras typically utilize infrared technology, which shares limitations with hardware-based methods. Moreover, the depth estimates provided by these methods do not precisely correspond to the world coordinates but rather represent scaled estimations (Li et al. 2021a) suspecting to scale ambiguity issues (Wang et al. 2018; Bian et al. 2019).

Many methods mentioned above have extensively utilized the Microsoft Kinect. The Kinect can be seen as a fusion of these methods and is recognized for its reliable localization accuracy, especially after scale adjustment. However, Kinect cameras can be expensive and have limited coverage (e.g., a maximum coverage area of $5 \times 5 \ m^2$ and perform optimally within a $3 \times 3 \ m^2$ space).

## Method

### Preliminary

Given a point $\mathbf{p}$ and its homogeneous world coordinate $\mathbf{p}_w = [\alpha, \beta, \gamma, 1]^\top$, we denote its projection as a pixel (homogeneous) coordinate in the image captured by a camera as $\mathbf{p}_c = [u, v, 1]^\top$. The relation of these two coordinates is

$$s \, \mathbf{p}_c = \mathbf{K} \, \mathbf{R} \, | \, \mathbf{T} \, \mathbf{p}_w = \mathbf{P} \, \mathbf{p}_w \qquad (3)$$

where $\mathbf{K} \in \mathbb{R}^{3 \times 3}$ denotes the matrix of intrinsic camera parameters, $\mathbf{R} \in \mathbb{R}^{3 \times 3}$ and $\mathbf{T} \in \mathbb{R}^{3 \times 1}$ are the extrinsic parameters defining the 3D rotation and 3D translation of the camera, and $s$ is a scale factor which is determined by the distance of $\mathbf{p}$ to the camera. The multiplication of $\mathbf{K}$, $\mathbf{R}$, and $\mathbf{T}$ yields a transformation matrix $\mathbf{P} \in \mathbb{R}^{3 \times 4}$ which combines the intrinsic and extrinsic parameters. The matrix $\mathbf{P}$ can be estimated using a set of points along with their corresponding world and pixel coordinates, known as the Perspective-n-Point (PnP) problem (Zheng et al. 2013; Hesch and Roumeliotis 2011). The Direct Linear Transformation (DLT), which replies on SVD to approximate the solution, is commonly used (Hartley and Zisserman 2003). Once $\mathbf{P}$ is estimated, it can be substituted into Eq. (3) for coordinate transformation. However, this transformation is only feasible from world coordinates to pixel coordinates (3D to 2D) because methods like DLT assume a single camera and cannot fully address the degree of freedom (DOF). To perform the reverse transformation (2D to 3D), multiple cameras are needed to ensure sufficient DOF. The Triangulation is adopted, which calculates the cross product of each side of Eq. (3) using a $\mathbf{p}_c$ as

$$\mathbf{p}_c \times s \, \mathbf{p}_c = \mathbf{p}_c \times \mathbf{P} \, \mathbf{p}_w, \qquad (4)$$

which leads to

$$\mathbf{A} \, \mathbf{p}_w = \mathbf{0}, \qquad (5)$$

where $\mathbf{A} = \mathbf{p}_c \times \mathbf{P}$ and the resulting $\mathbf{0}$ is obtained because $\mathbf{p}_c \times \mathbf{p}_c = \mathbf{0}$. Eq. (5) can be utilized to estimate the $\mathbf{p}_w$ using a given set of points and their pixel coordinates in multiple cameras. Once again, the SVD method needs to be utilized.

### Rethink From A Probabilistic Perspective

Let us reformulate the problem from a probabilistic standpoint. The objective is to predict the world coordinate (denoted by a random variable $E[\mathbf{Y}] = \mathbf{p}_w \in \mathbb{R}^{4 \times 1}$) of a point $\mathbf{p}$ using its corresponding pixel coordinates (denoted by a random variable $E[\mathbf{X}] = [\mathbf{p}_c] \in \mathbb{R}^{3 \times k}$ where $k$ is the number of cameras). This prediction is based on a training dataset **D** consisting of samples drawn from the distributions of the **Y** and **X**. We can write it as a mapping function $\hat{Y} = g(X; D)$. The expectation of this function is

$$E[\hat{Y}|X; D] = \int_{x \sim X} g(x; D) F_{\hat{Y}|X,D}(g(x)|x, D) dx, \quad (6)$$

where the $F_{\hat{Y}|X,D}(\cdot)$ is the conditional probability density function (PDF) of $\hat{Y}$ given the observation of **X** and **D**. In traditional methods using the PnP and Triangulation, the expectation of the function is not easy to converge to the expectation of **Y** (i.e., $E[Y]$) for three reasons. Firstly, only a single sample $x$ can be drawn for each **X** at a specific point **p**, resulting in the expectation being equivalent to $E[\hat{Y}|X; D] = g(x; D) F_{\hat{Y}|X,D}(g(x)|x, D)$. This is considerably less efficient as a statistic for estimating the true expectation. Secondly, the probability distribution $F_{\hat{Y}|X,D}(\cdot)$ is likely skewed and leads to the biased predictions. Lastly, The training dataset **D** has a limited size, typically a few hundred samples, which reduces the likelihood of the learned model $g(X; D)$ producing globally optimal solutions. Our method has addressed those issues.

### Expanding the Sampling Scope

In tasks related to human localization, there is no need to model every individual point. Instead, the target point **p** represents the geometric center of the body. We can regard all points on the human body as the observations of the center. Consequently, we can sample a large number of world (or pixel) coordinates $Y_i$ (or $X_i$), which are theoretically drawn from the distribution of **Y** (or **X**) (see Fig. 1). For $X_i$'s, we can gather observations by sampling multiple batches of points from the resulting bounding box of pedestrian detection at each camera (Sun et al. 2019a).

### Mitigating Distribution Biases

To mitigate the distribution biases caused by the skewness of the distributions, we design mean estimators as

$$\overline{X} = \frac{1}{m} \sum_{i=1}^{m} X_i, \quad \overline{Y} = \frac{1}{m} \sum_{i=1}^{m} Y_i \quad (7)$$

where the $1 \leq m \leq n$ is the number of original observations selected to construct the mean estimator and $n$ is the total number of $Y_i$ or $X_i$ available. Note that we can sample multiple batches of points for both mean estimators by varying the $m$ to obtain subsets of $Y_i$ or $X_i$. The maximum number of mean estimators thus can reach $(2^n - 1)^2$. With 20 original observations, this can generate 1,099,509,530,625 $(\overline{X}, \overline{Y})$ pairs. An illustration is shown in Fig. 2. We denote the new training dataset of mean estimators as $\overline{D}$ hereafter. The relaxation in pairing enables a much flexible sampling which allows us to conduct any learning sufficiently.

The rational behind this flexibility is that the expectation of random variables from the same distribution is equal to that of the mean of those variables (i.e., $E[X_i] = E[X] = E[\overline{X}]$ and $E[\overline{Y_i}] = E[\overline{Y}] = E[Y]$). Additionally, the Law of

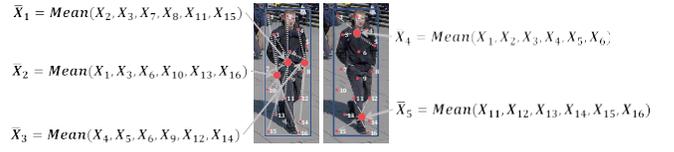

Figure 2: Illustration of sampling 5 batches of mean estimators from the bounding box.

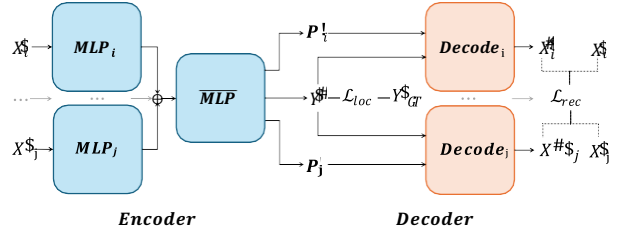

Figure 3: Neural implementation of MoM with encoder-decoder collaboration.

Large Numbers (LLN) states that the average of them converges to the true value as

$$P \lim_{n \to \infty} \overline{X} = \overline{p}_c, \quad P \lim_{n \to \infty} \overline{Y} = \overline{p}_w. \quad (8)$$

Such a large number of samples also ensures $\overline{X}$, $\overline{Y}$, and $g(\overline{X}; \overline{D})$ all follow a normal distribution as

$$\overline{X} \sim N(\overline{p}_c, \Sigma_{\overline{X}}), \overline{Y} \sim N(\overline{p}_w, \Sigma_{\overline{Y}}), g(\overline{X}; \overline{D}) \sim N(\overline{p}_w, \Sigma_g) \quad (9)$$

where the $\Sigma$'s are the covariance of corresponding variables. This assurance is derived from the Central Limit Theorem (CLT), which states that the sampling distribution of the mean will always follow a normal distribution, provided that the sample size is sufficiently large. This helps to address the distribution bias issue in Eq. (6).

### Improving Prediction Accuracy using Autoencoder

Regarding prediction accuracy, we propose replacing the strict SVD solvers with a more flexible autoencoder learner shown in Fig. 3. The encoder is a neural network implementation of the mapping function $g(\cdot)$ (Eq. (6)) which predicts the mean of world coordinate from the corresponding pixel coordinates of $k$ cameras. On the other hand, the decoder is responsible for constraining the predictions to adhere to the classical perspective model described in Eq. (3), thereby reducing the risk of the neural encoder learning an overfitted solution. The design principles are as follows.

**The Mean of Means Encoder**: The encoder is a $k$-stream MLP which implements the function $g(\cdot)$ with the intention to combine the PnP and Triangulation into an end-to-end network. As discussed early, we conduct the learning using the extensive sampling of mean pairs of $(\overline{X}, \overline{Y})$. As shown in Eq. (2), the function to learn can be rewritten as $\overline{\hat{Y}} = \overline{g}(\overline{X}; \overline{D})$. We refer to this model as Mean of Means (MoM) since it predicts the mean of the world coordinates using mean estimators of pixel coordinates. MoM enhances prediction accuracy from three perspectives.

Firstly, this end-to-end architecture reduces the risk of error propagation in traditional multi-stage SVD solvers and increases the likelihood of obtaining a globally optimal model. Each stream of the MLP corresponds to a specific camera. It takes from each camera a set of mean estimations and feed them into a local MLP (denoted as $\mathbf{MLP}_k$) for camera-specific parameter learning. The features of all streams are then concatenated and fed into a global MLP (denoted as $\mathbf{MLP}$) for world coordinate prediction. The neural implemented of the encoder is then written

$$\bar{g}(\bar{\mathbf{X}}; \bar{\mathbf{D}}) = \overline{\mathbf{MLP}}\left(\bigoplus_k \overline{\mathbf{MLP}_k}\left(\bar{\mathbf{X}}_k\right)\right), \quad (10)$$

where $\oplus$ is for feature concatenation. Note we set the size of mean estimator set to 12 (i.e., $\bar{\mathbf{X}}_k \in \mathbb{R}^{12 \times 2}$) in the sense that in traditional PnP, there are 12 DOF. The MLPs consist of fully connected layers (FCs) and use ReLU as the activation function. More details can be found in our source code. To conduct the learning for the encoder, we employ L2 norm as the loss function for the encoder

$$L_{loc} = \sum_{\bar{x} \sim \bar{\mathbf{X}}} \left\| \bar{g}(\bar{\mathbf{x}}; \bar{\mathbf{D}}) - \bar{y} \right\|, \quad (11)$$

where every training pair $(\bar{x}, \bar{y}) \in \bar{\mathbf{D}}$ is sampled from the possible mean estimator pairs $(\bar{\mathbf{X}}, \bar{\mathbf{Y}})$.

The design of this encoder is inspired by the findings in (Brutzkus and Globerson 2017), where the authors discovered that a CNN with non-overlapping convolutions and ReLU activations can guarantee a globally optimized solution when the inputs follow a normal distribution. Our model satisfies the required conditions and is expected to generate predictions with higher accuracy.

Secondly, our implementation of predicating the mean using mean estimators has indeed transferred the problem of predicting the single-point coordinate into a problem of estimating the expectation of the coordinate. Furthermore, our extensive sampling ensures we have a comprehensive coverage of the supports of the $\bar{\mathbf{X}}$, $\bar{\mathbf{Y}}$, and $\bar{\mathbf{D}}$. By the Law of Iterated Expectations (LIE), the expectation of the function $E(\bar{g}(\bar{\mathbf{X}}; \bar{\mathbf{D}}))$ converges to the $E(\mathbf{Y})$. This can be proven as

$$\begin{aligned} E\left[\bar{g}(\bar{\mathbf{X}}; \bar{\mathbf{D}})\right] &= E\left[E\left[\hat{\bar{\mathbf{Y}}} | \bar{\mathbf{X}}, \bar{\mathbf{D}}\right]\right] \\ &= \int_{\bar{x} \sim \bar{\mathbf{X}}} \int_{\bar{d} \sim \bar{\mathbf{D}}} E\left[\hat{\bar{\mathbf{Y}}} | \bar{x}, \bar{d}\right] P(\bar{x}, \bar{d}) \, d\bar{d} \, d\bar{x} \quad \text{(by LOTUS)} \\ &= \int_{\bar{x} \sim \bar{\mathbf{X}}} \int_{\bar{d} \sim \bar{\mathbf{D}}} \int_{\hat{\bar{y}} \sim \hat{\bar{\mathbf{Y}}}} \hat{\bar{y}} \, P(\hat{\bar{y}} | \bar{x}, \bar{d}) \, d\hat{\bar{y}} \, P(\bar{x}, \bar{d}) \, d\bar{d} \, d\bar{x} \\ &= \int_{\bar{x} \sim \bar{\mathbf{X}}} \int_{\bar{d} \sim \bar{\mathbf{D}}} \int_{\hat{\bar{y}} \sim \hat{\bar{\mathbf{Y}}}} \bar{y} \, P(\hat{\bar{y}}) \, d\hat{\bar{y}} \, d\bar{d} \, d\bar{x} \\ &= E[\hat{\bar{\mathbf{Y}}}] = E[\bar{\mathbf{Y}}] = E[\mathbf{Y}] \quad \text{(by Eq. (9) and (11))} \end{aligned} \quad (12)$$

where LOTUS is the Law of the Unconscious Statistician.

Lastly, the use of mean estimators and the assurance of normality not only increase the flexibility of sampling and learning, but also enhance the model's resistance to noise and perturbations by allowing proper cancellation of positive and negative biases. In the experiments, we will demonstrate the robustness of the model against camera motions and various types of noise.

**Perspective Transformation Regulated Decoder**: Neural networks show promise as function learners. However, they often face the issue of overfitting, especially when dealing with a large number of parameters compared to traditional methods (e.g., only 12 parameters in the transformation matrix $\mathbf{P}$). To address this, we introduce a decoder that works in tandem with the MoM encoder. The concept is that if the encoder accurately predicts a world coordinate $\mathbf{Y}$ for a point $\mathbf{p}$, then the predicted $\hat{\mathbf{Y}}$ should be convertible back to $\mathbf{p}$'s pixel coordinate $\bar{\mathbf{X}}$ using a traditional perspective transformation (e.g., Eq. (3)). Since the theory behind traditional perspective transformation is well-established, the decoding process encourages the encoding learner to avoid learning an overly complex function. This enables validation of predictions and regulates the learning process.

To achieve this, we extend the MoM encoder's output with $k$ transformation matrices $\bar{\mathbf{P}}_k$. It is important to note that these $\bar{\mathbf{P}}_k$ serve as analogies for $\mathbf{P}$ in Eq. (3). In traditional SVD solvers, the factor $s$ is cancelled during the calculation, which is one reason that accurate world coordinates cannot be estimated. In an end-to-end neural implementation, we can overcome this limitation and learn the factor $s$ along with $\mathbf{P}$. Each transformation matrix $\bar{\mathbf{P}}_k$ is essentially an estimation of $\frac{\mathbf{P}}{s}$. The decoder for regulation is then a $k$-stream MLP where in each stream, the calculation is equivalent to traditional perspective transformation of

$$\hat{\bar{\mathbf{X}}} = \mathbf{Decode}_k\left(\hat{\bar{\mathbf{Y}}}\right) = \bar{\mathbf{P}}_k \hat{\bar{\mathbf{Y}}}. \quad (13)$$

The loss for the decoder is written

$$L_{rec} = \sum_k \left\| \mathbf{Decode}_k\left(\hat{\bar{\mathbf{Y}}}\right) - \bar{\mathbf{X}}_k \right\|, \quad (14)$$

where $\bar{\mathbf{X}}_k$ is the mean of pixel observations of the $k^{th}$ camera. The encoder and decoder work together to form an autoencoder, collectively addressing the learning issues.

# Experiment

We implement MoM using PyTorch and conduct experiments on a system equipped with 64 CPU cores, 96 GB RAM, and an NVIDIA RTX 4090 GPU. The implementation details and dataset can be found in source code.

## Datasets

We evaluate the method using 4 datasets obtained from environments spanning both indoor and outdoor settings, covering an area up to $12 \times 36 \, m^2$.

**Indoor Walking Dataset** We construct a human localization benchmark in a real-world indoor environment. The dataset consists of over 34,160 samples of human location in an indoor space of $10 \times 10 m^2$. We use two web cameras (MF-100) to record videos (640×480 pixels) of three distinct walking patterns: random walking within the space, walking along a predefined path with a cross shape (cross walk), and walking along a predefined square path measuring (square walk). Besides, we also use a Kinect (Azure Kinect DK) as well as a UWB system (BP-TWR-50) to capture world coordinates. Fifteen subjects (seven males and eight females) participate in the data collection. The ground truth of the human locations are initially obtained using Kinect and then corrected by aligning to a set of manually annotated locations. The 2D human keypoints are extracted by HRNet (Wang et al. 2020), while the bounding boxes are detected by YOLOv8 (Jocher, Chaurasia, and Qiu 2023).

Table 1: Performance comparison on human localization on four datasets. The best results are in bold font.

| Dataset | Methods | Position Error (m) | | | Traj. Error (m) | | Accuracy at Different Thresholds (%) | | | |
|---|---|---|---|---|---|---|---|---|---|---|
| | | Mean ↓ | Median ↓ | Std. ↓ | ATE ↓ | RPE ↓ | @0.2m ↑ | @0.3m ↑ | @0.4m ↑ | @0.5m ↑ |
| Walking (Indoor) | PnP + Triangulation | 0.376 | 0.261 | 0.758 | 0.512 | 0.202 | 33.428 | 60.223 | 81.429 | 93.650 |
| | UWB | 0.261 | 0.257 | 0.116 | 0.326 | 0.331 | 32.231 | 63.407 | 87.602 | 97.782 |
| | MoM (ours) + keypoint | 0.152 | 0.135 | 0.098 | 0.296 | 0.090 | 78.537 | 94.465 | 98.393 | 99.216 |
| | **MoM (ours) + bbox** | **0.142** | **0.132** | **0.075** | **0.201** | **0.072** | **81.532** | **96.975** | **99.575** | **99.962** |
| AIST++ (Indoor) | PnP + Triangulation | **0.073** | **0.040** | 0.082 | **0.075** | **0.008** | 90.331 | 99.764 | 99.855 | 99.885 |
| | **MoM (ours) + keypoint** | 0.075 | 0.057 | **0.062** | 0.082 | 0.030 | **97.117** | **99.774** | **99.864** | **99.907** |
| | MoM (ours) + bbox | 0.116 | 0.101 | 0.072 | 0.144 | 0.040 | 88.700 | 98.241 | 99.584 | 99.834 |
| Wildtrack (Outdoor) | PnP + Triangulation | 0.393 | 0.316 | 0.319 | 0.303 | 0.419 | 34.068 | 43.819 | 69.414 | 73.709 |
| | MoM (ours) + keypoint | 0.160 | 0.120 | 0.140 | 0.158 | 0.232 | 75.084 | 88.215 | 93.660 | 96.521 |
| | **MoM (ours) + bbox** | **0.078** | **0.067** | **0.055** | **0.139** | **0.111** | **96.169** | **99.478** | **99.942** | **99.942** |
| MultiviewX (Outdoor) | PnP + Triangulation | 15.617 | 4.032 | 100.368 | 15.094 | 26.750 | 1.654 | 3.488 | 5.090 | 7.003 |
| | MoM (ours) + keypoint | 0.130 | 0.111 | 0.088 | 0.295 | 0.185 | 82.196 | 95.013 | 98.760 | 99.690 |
| | **MoM (ours) + bbox** | **0.113** | **0.096** | **0.075** | **0.145** | **0.159** | **88.140** | **97.131** | **99.483** | **99.974** |

**AIST++ Dance Dataset (Li et al. 2021b)** This is an indoor dataset for dance performance conducted within a circular area with a radius of 5 meters. There are 35 dancers and 10 dance motion genres, which provides a rich diversity in terms of persons and action motions. The dataset provides 2D human keypoints on 9 cameras (1920×1080 pixels) and 3D human keypoint annotated using multiview capture and manually initialized camera parameters. The training set includes 20 subjects and 712,919 images, while the testing set contains 10 subjects and 352,948 images.

**Wildtrack Dataset (Chavdarova et al. 2018)** There are 313 individuals standing and walking captured by 7 cameras (1920×1080 pixels) in a 12×36$m^2$ open area. The annotated human positions are derived from the ground plane, which is quantified into a 480×1440 grid, where each grid cell measures 2.5$cm$ squared. The training split consists of 3,941 samples, and testing split includes 1,723 samples.

**MultiviewX Dataset (Hou, Zheng, and Gould 2020)** This is a synthetic dataset built using 350 pedestrian models and scenario created in Unity. There are 6 cameras (1920×1080 pixels) with overlapping field-of-view and it covers a square of 16×25$m^2$. The training subset includes 8,585 samples, and testing subset includes 3,870 samples.

### Evaluation Metrics

We use several metrics to evaluate the performance, including 1) Positioning Errors: Mean, Median and Stan-

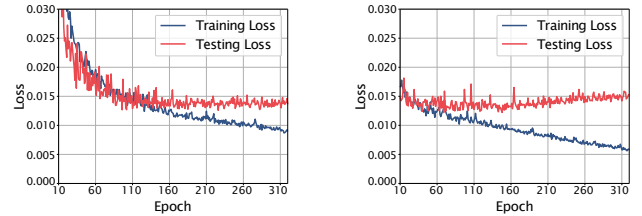

(a) With decoder.  (b) Without decoder.

Figure 4: Training and testing losses of MoM.

dard deviation; 2) Localization Accuracy: we use several thresholds, i.e., Acc@0.2m, @0.3m, @0.4m, @0.5m, where Acc@0.2m is the accuracy that a prediction is counted as accurate if the distance between the predicted location and the ground truth is less than 0.2 $m$; and 3) Trajectory Errors: absolute trajectory error (ATE) and relative pose error (RPE).

### Comparison with SOTA Methods

We compare MoM with the PnP+Triangulation solution and a hardware-based method (i.e., UWB) (Cheng and Zhou 2019) as the representatives of the commonly used stereo vision and hardware methods. The results are shown in Table 1. MoM achieves an accuracy exceeding 96% across all datasets at a 0.3m threshold. It outperforms competing methods on every metric within the Wildtrack, MultiviewX, and our Indoor Walking datasets. Even on the AIST++ dataset, where MoM leads in accuracy across all thresholds, it only slightly trails traditional methods in some metrics, with minimal discrepancies of just 0.2$cm$ to 2.2$cm$. Surprisingly, we discovered that the mean error for the Kinect device within and beyond its optimal performance range of 5×5$m^2$ is 0.021$m$ and 0.136$m$, respectively. On the other hand, MoM achieves a mean error of 0.033$m$ and 0.116$m$, indicating superior localization capabilities over greater distances. In terms of efficiency, MoM achieves a real-time inference speed of 1,060 samples per second on the CPU.

### Collaboration of the Encoder and the Decoder

We compare the performances of the MoM model with and without the decoder in Table 2. With the help of the decoder,

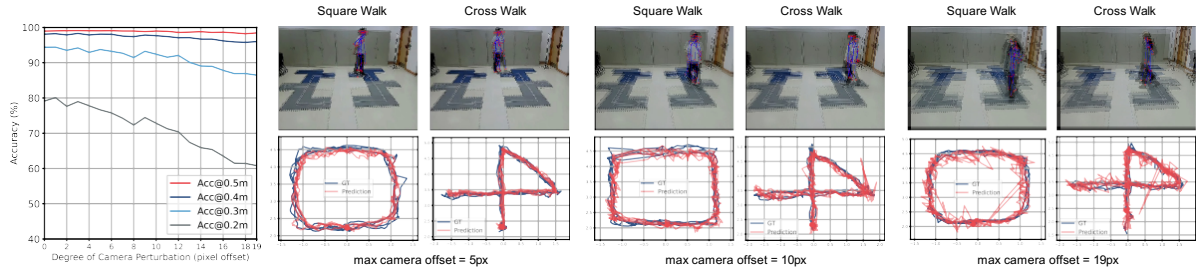

Figure 5: The performance of MoM was evaluated under varying degrees of camera perturbation with the predicted trajectories compared with the ground truth (GT): no significant drops in performance within the range of 0.3m were observed until the maximum camera offsets exceeded 12 pixels.

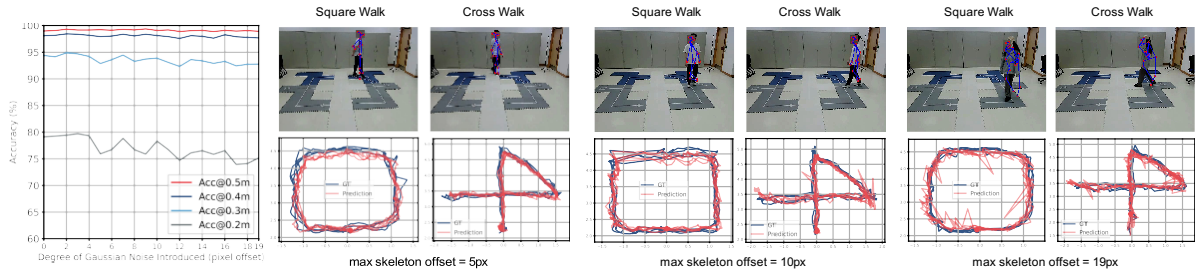

Figure 6: The performance of MoM over various degrees of noise introduced into skeleton detection: the keypoints on the skeleton have been deteriorated with offsets in pixels. No significant drops in performance were observed within all ranges.

Table 2: MoM performance over the decoder regulation.

| Decoder Settings | Accuracy at Different Thresholds (%) | | | |
|---|---|---|---|---|
| | @0.2m ↑ | @0.3m ↑ | @0.4m ↑ | @0.5m ↑ |
| w/o | 76.254 | 93.674 | 97.982 | 98.992 |
| w/ | **78.537** | **94.465** | **98.393** | **99.216** |

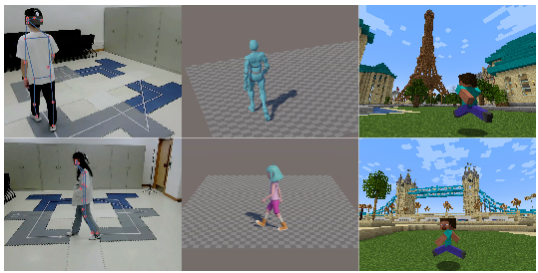

Figure 7: Applications by integrating MoM with Avatars in Unity and Minecraft.

the encoder achieves higher accuracy at all thresholds. The effectiveness of the collaboration of the encoder and decoder is further verified by the training and testing losses in Fig. 4. With the decoder, the testing loss converges to 0.013 after 160 epochs, superior to that without the decoder regulation which 310 epoch to converge to 0.015.

### Performance Over Camera Motions

We verify the effectiveness of our method against different degrees of camera perturbation. Specifically, we stimulate the perturbation by adding random pixel offsets to the input videos. Fig. 5 shows the performance over the pixel offset from 0 to 19. It is clear that there are no significant changes on Acc@0.3m when the offset is less than 12.

### Performance Over Noise

We also verify the robustness of MoM over different degrees of Gaussian noise. Specifically, the detected joint points are shifted to a random $n$ point of the Gaussian distribution to build up a new noisy skeleton. Fig. 6 compares the performance with $n$ ranging from 0 to 19. There are no significant differences on Acc@0.4m and 0.5m with different noise values. When $n = 19$, Acc@0.3m drops slightly by 2%.

### Supplementary Studies and Applications

To study the practicality of the MoM, we have also investigated the impacts of dataset splitting, number of cameras, and walking patterns. However, due to the space limitations, the results and analysis have been included in the supplementary materials. In addition, we have built two applications by successfully integrating MoM with two avatars. The first avatar is a walking robot implemented in Unity, while the second avatar is Steve from Minecraft. Several examples of applications can be seen in Fig. 7. For additional videos and examples, please refer to the supplementary materials.

## Conclusion

This paper introduces a unified framework called Mean of Means (MoM) to address challenges related to one-to-one pairing of world and pixel coordinates in human body localization. MoM relaxes this requirement by considering the entire body as a distribution generated from the mean or geometric center. The advantages of MoM include large-scale sampling, normality consistency, and a balance between neural implementation and classical theory. Large-scale sampling allows flexibility in data collection, enabling the estimation of a single center using sampled points from the body. Normality consistency models the relation between mean estimators, ensuring that the means of body and pixel coordinates follow normal distributions. The balance between neural implementation and classical theory is achieved through an end-to-end autoencoder framework, replacing multi-stage solvers with neural mapping functions and leveraging a decoder for validation. The performance of MoM has been validated in the experiments.